\documentclass{article}

\usepackage{mlapa}
\input{Definitions}
\title{Bayesian two-sample tests}
\author{Karsten M. Borgwardt$^1$ and Zoubin Ghahramani$^2$ \\ $^1$Max-Planck-Institutes T\"ubingen, $^2$University of Cambridge}

\begin{document}
\maketitle
%\twocolumn[
%\icmltitle{Bayesian two-sample tests}
%\icmlauthor{}{}
%\icmladdress{{\bf Keywords}: two-sample problem, Bayesian hypothesis testing, Dirichlet process mixture models }
%\vskip 0.3in
%]
%

\begin{abstract}
In this paper, we present two classes of Bayesian approaches to the two-sample problem.
Our first class of methods extends the Bayesian t-test to include all parametric models in the exponential family and their conjugate priors. Our second class of methods uses Dirichlet process mixtures (DPM) of such conjugate-exponential distributions as flexible nonparametric priors over the unknown distributions. %On synthetic examples and real medical datasets, we show that our tests are competitive with the best state-of-the-art methods for this task, even outperforming them on average on the medical datasets.
%Judging whether two sets of observations were generated by the same underlying process is a fundamental statistical question, with important applications to the life  sciences and the social sciences. In statistics, this is known as the two-sample problem: given two data sets, decide whether they were generated from the same distribution or from different distributions. While this problem has been addressed in classical statistics and machine learning, we present here the first two Bayesian solutions to the two-sample problem and compare them to existing methods. 
%Our first class of methods extends the Bayesian t-test to include all parametric models in the exponential family and their conjugate priors. Our second class of methods uses Dirichlet process mixtures (DPM) of such conjugate-exponential distributions as flexible nonparametric priors over the unknown distributions. 
%Our test statistic is the Bayes factor comparing the marginal likelihood of the hypothesis that the samples were generated from one unknown distribution versus two. An algorithm based on hierarchical clustering is used to approximate these marginal likelihoods efficiently for the DPM.
%We experimentally evaluate the  performance of our novel two-sample tests and compare them to ground truth and to state-of-the-art two-sample tests from classic statistics and current machine  learning.
\end{abstract}
\newtheorem{problemstatement}{Problem Statement}
\newcommand{\D}{{\mathcal D}}
\newcommand{\bx}{{\mathbf x}}
\newcommand{\bu}{{\mathbf u}}
\newcommand{\bnu}{{\mathbf \nu}}
\newcommand{\alphat}{\tilde{\alpha}}
\newcommand{\betat}{\tilde{\beta}}
\newcommand{\ttt}{\raisebox{0.6ex}{\scriptsize \texttrademark}}

\section{Introduction}
\label{sec:introduction}

In this paper, we tackle the so-called two-sample problem:
\begin{problemstatement}
  Given two samples $X=\{x_1,\ldots,x_{m_1}\} \sim q_1$ and $Y=\{y_1,\ldots,y_{m_2}\} \sim q_2$ from two underlying distributions $q_1$ and $q_2$. The two-sample problem is to decide whether $q_1=q_2$.
\end{problemstatement}
An associated test is called a two-sample test. Such tests are encountered in various disciplines from the life sciences to the social sciences:
\begin{itemize}
\item In medical studies, one may want to find out if two classes of patients show different behaviour, response to a drug or susceptibility to a disease.
\item In microarray analysis, one may compare measurements from different weeks, labs or platforms to find out if they follow the same distribution, before integrating them into one dataset, in order to increase sample size.
\item In the neurosciences, one may want to compare measurements of brain signals under different external stimuli, to check whether brain activity is affected by these stimuli.
\item In the social sciences, one may want to compare whether the behavior of a group of people, e.g. when they graduate, marry, or die, is different across countries or generations.
\item In the financial sciences, one could for example compare the set of transactions performed at a stock exchange during different weeks, to find out if there is a change in activity in the financial markets. 
\end{itemize}

While this question has been studied in detail by classic statistics for univariate data, there is less work on multivariate data (which we review in Section \ref{sec:backgr-two-sample}). The only machine learning approach to this problem is a kernel method by \cite{GreBorRasSchSmo07}, using the means of the two samples $X$ and $Y$ in a universal reproducing kernel Hilbert Space as its test statistics, but it has created lots of interest in that subject and follow-on studies~\cite{BorGreRasKriSchSmo06,HuaSmoGreBorSch07,GreGyo08}.

Here, we approach this two-sample problem from a Bayesian perspective. The classic Bayesian formulation of this problem would be in terms of a Bayes factor~\cite{KasRaf95} which represents the likelihood ratio that the data were generated according to hypothesis $\Hcal_0$ (that is from the same distribution) or hypothesis $\Hcal_1$ (that is from different distributions). However, how to exactly define these two hypotheses is a crucial question, and many answers have been given in the Bayesian literature with hypotheses that are tailored to a specific problem or application domain; one example are the Bayesian t-tests used in microarray data analysis~\cite{baldi_bayesian_2001,fox_two-sample_2006}. Our goal in this paper is to define two general classes of two-sample tests that represent a precise formulation of the two-sample problem, but are not tailored to a specific application. They are designed to offer an attractive middle ground between the general idea of using Bayes factors and the specialised hypotheses testing problems studied in the literature.

In detail, we define a class of nonparametric Bayesian two sample tests based on Dirichlet process mixture models. 
The use of Dirichlet process mixtures for flexible nonparametric modelling of general unknown distributions has a long history in Statistics. However, although the two-sample problem depends crucially on testing whether data come from one or two unknown distributions, Bayesian approaches based on nonparametric density models have not been explored to date. Here we propose and explore such a non-parametric method using the classic Dirichlet process mixture.
To the best of our knowledge, the only work that is remotely related is that on a Bayesian test for a parametric versus a nonparametric model of the data by Berger and Guglielmi~\cite{berger_bayesian_1998}. This addresses a different but related question since it assumes a parametric null hypothesis. We also define a parametric Bayesian two-sample test where the model of the data is a member of the exponential family. This test generalizes the Bayesian t-test by~\cite{baldi_bayesian_2001} and~\cite{fox_two-sample_2006}, who assume that the samples are Gaussian.

%Bayesian in nature, it possesses all the attractive properties of a Bayesian test.

% XXX update
This paper is structured as follows. In Section~\ref{sec:backgr-two-sample} we will review existing approaches to the two-sample problem on multivariate data, and highlight some differences between frequentist and Bayesian hypothesis testing.
In Section~\ref{sec:bayesian-two-sample} we outline the common core of our two Bayesian two-sample test, before providing the details on the parametric test in Section~\ref{sec:param-bayes-two} and on the non-parametric  test in Section~\ref{sec:nonp-bayes-two}. %In Section~\ref{sec:experiments} we will evaluate the performance of our tests, comparing it to the state-of-the-art methods in this field.

%\begin{itemize}
%\item 
%Advantages:
%Bayesian, can include prior knowledge, no need to pick kernel and its bandwidth, faster(?), better(?)
%
%\item Related Work:
%Classic statistics, MMD, Jim Berger, bayesian t-test from bioinformatics.
%\end{itemize}

\section{Multivariate two-sample tests}
\label{sec:backgr-two-sample}

\paragraph{Related work in statistics and kernel machines} Our method is a Bayesian approach to a problem that has been studied in classic statistics and kernel machine learning. Here we describe in short the prominent multivariate two-sample tests (see also~\cite{GreBorRasSchSmo07}).

Frequentist two sample tests follow the same principle of classic hypothesis testing: Given the two samples $X$ and $Y$, a test statistic is computed. Then the distribution of this test statistics under the null distribution ($q_1=q_2$) is determined. If the value of the test statistic falls into the $1-\alpha$-quantile of the null distribution, the null hypothesis $q_1=q_2$ is accepted at significance level $\alpha$. If its value exceeds the $1-\alpha$ quantile, it is rejected at significance level $\alpha$. So the outcome of these test depends on the significance level $\alpha$ which has to be chosen apriori.

Frequentists tests differ mainly in two points: a) the test statistic they employ and b) the way in which they determine the null distribution for this test statistic. 
The classic \textit{multivariate t-test}~\cite{hotelling51} assumes that both distributions are multivariate Gaussian with unknown identical covariance;
\textit{Friedman and Rafsky}~\cite{FriRaf79,HenPen99} define test statistics based on spanning trees, namely the number of edges that connect points from $X$ to $Y$ in a minimum spanning tree (Wald-Wolfowitz test) and the closeness of points from $X$ and $Y$ in a ranking derived from the minimum spanning tree (Kolmogorov-Smirnow test)~\cite{Bic69,FriRaf79}. \textit{Rosenbaum}'s test statistic is the number of pairs containing a data point from $X$ and $Y$ in a minimum distance non-bipartite matching over $X \cup Y$. \textit{Hall and Tajvidi}~\cite{HalTaj02} essentially for each data point count its number of nearest neighbours in $X \cup Y$ that are from the other sample. \textit{Biau and Gyorfi}'s statistic is the distance between Parzen window estimates of the densities~\cite{AndHalTit94,BiaGyo05}. \textit{Gretton et al.} use the distance between the means of $X$ and $Y$ in a universal reproducing kernel Hilbert space as their test statistic~\cite{GreBorRasSchSmo07}.

\paragraph{Frequentist versus Bayesian approach}

In contrast to classic hypothesis testing, the test statistic in Bayesian hypothesis testing is a so-called {Bayes factor}. It is the ratio of the likelihoods of two opposing hypotheses having generated the data $D = \{X,Y\}$, the hypothesis $\Hcal_0$ ($q_1=q_2$) and its alternative $\Hcal_1$ ($q_1 \ne q_2$).
 
To summarize, frequentist classic hypothesis testing considers only one hypothesis and evidence {\it against} it, whereas Bayesian hypothesis testing compares the likelihoods of two alternative hypotheses having generated the data at hand. While the question of which perspective is to prefer is still an ongoing and unresolved debate, we deem it useful to have a Bayesian alternative to the classic frequentist two sample tests for the following reasons: Bayesian approaches have a clear interpretability compared to the commonly used p-values.
Prior knowledge on the probability of the two hypotheses can be incorporated into the Bayes factor in a straightforward manner.

\section{Concept of Bayesian two-sample tests}
\label{sec:bayesian-two-sample}

\subsection{Bayes factor as test criterion}
\label{sec:bayes-factor-as}
%ratio of log likelihoods of H1 and H0
Our two classes of Bayesian two sample tests are based on the idea to compute a Bayes factor between two alternative hypotheses: the hypothesis $\Hcal_1$ that both samples were independently generated from different underlying distributions $q_1$ and $q_2$ with $q_1 \ne q_2$, and the hypothesis $\Hcal_0$ that they originated from the same distribution $q$ ($q_1=q_2$). This idea is formalised in the following lemma. 

\begin{lemma} Given two samples $X \sim q_1$ and $Y \sim q_2$, we accept the hypothesis $\Hcal_1$ that $q_1 \ne q_2$ if the Bayes factor
\begin{align}  
\chi =  \frac{P(X,Y| \Hcal_1)}{P(X,Y| \Hcal_0)} > 1, \label{eq:6}
\end{align}  
otherwise we accept the hypothesis $\Hcal_0$ that $q_1=q_2=q$.
\end{lemma}

The hypothesis $\Hcal_1$ is that the samples originate from different distributions, such that \begin{align} P(X,Y|\Hcal_1) = P(X|\Hcal_1) P(Y|\Hcal_1).\end{align}

\subsection{Computation of the Bayes factor}
\label{sec:comp-bayes-fact}
The central challenge when computing the Bayes factor $\chi$ is that we do not know the distributions $q_1$, $q_2$, and $q$ our Bayes factor is based upon. Since $q$, $q_1$ and $q_2$ are unknown probability distributions, we have to compute the integral over all such distributions with respect to some prior on distributions. 
We offer two classes of solutions here. In Section~\ref{sec:param-bayes-two}, we present a parametric test  where the distributions are in the exponential family and have conjugate priors. In Section~\ref{sec:nonp-bayes-two}, we present a non-parametric test where the distributions $q_1$,$q_2$, and $q$ are assumed to be drawn from a Dirichlet Process mixture model.

\section{Parametric Bayesian two-sample test}
\label{sec:param-bayes-two}
\subsection{Exponential Families}
\label{sec:exponential-families}

For the parametric Bayesian two-sample test, we assume that the underlying distributions $q_1$ and $q_2$ are in the exponential family:
The distribution for models from this family can be written in the form
\begin{align} p(\xb|\theta) = f(\xb) g(\theta) \exp \{ \theta^\top \bu(\xb) \}, \end{align}
where $\bu(\xb)$ is a $K$-dimensional vector of sufficient statistics,
$\theta$ are the natural parameters, and $f$ and $g$ are non-negative
functions.  The conjugate prior is 
\begin{align} p(\theta|\eta,\bnu) = h(\eta,\bnu) g(\theta)^\eta \exp \{ \theta^\top
\bnu \}, \end{align}
where $\eta$ and $\bnu$ are hyperparameters, and $h$ 
normalizes the distribution.

\subsection{Bayes factor of parametric test}
\label{sec:bayes-fact-param}

The Bayes factor of the parametric two-sample test can then be computed as 
\begin{align}
  \label{eq:11}
 \chi &=  \frac{P(X|\beta) P(Y|\beta)}{P(X,Y|\beta)} = \\
      &=  \frac{\int P(X|\theta) P(\theta | \beta) d \theta \int P(Y|\theta) P(\theta | \beta) d \theta }{\int P(X,Y|\theta) P(\theta | \beta) d \theta} = \\
      &= \frac{h(\eta,\bnu) \, h(\eta+m_1+m_2, \bnu+\ub(X) + \ub(Y))}{h(\eta + m_1, \nu + \ub(X)) \,h(\eta + m_2, \nu + \ub(Y)) },
\
\end{align}
where \begin{align}\ub(X) = \sum_{i=1}^{m_1} \ub(x_i), \ub(Y) = \sum_{j=1}^{m_2} \ub(y_j), \end{align} and $\beta$ is the set of hyperparameters $\{\eta,\bnu\}$ of the prior.

\section{Nonparametric Bayesian two-sample test}
\label{sec:nonp-bayes-two}
Unlike its parametric counterpart, our nonparametric Bayesian two-sample test does not employ one single model for the data, but rather the limit of infinitely many components of a finite mixture model: $P(X|\alpha,\beta) = \int P(X|q) P(q|\alpha,\beta) d q$ where $q$ is an unknown distribution, modelled as an infinite mixture,  and $\alpha$ and $\beta$ are hyperparameters controlling it. 

This can be achieved via a Dirichlet process mixture of members of the exponential family. The Bayes factor for the nonparametric two-sample test equals
\begin{align}
  \label{eq:12}
  \chi &= \frac{P(X|\alpha,\beta)P(Y|\alpha,\beta)}{P(X,Y|\alpha,\beta)}
%       &=  \frac{\int P(X|\theta) P(\theta | \alpha, \beta) d \theta \int P(Y|\theta) P(\theta | \alpha, \beta) d \theta }{\int P(X,Y|\theta) P(\theta | \alpha, \beta) d \theta},
\end{align}
where $P(X|\alpha,\beta)$ is the marginal probability of sample $X$ under a Dirichlet Process Mixture Model (analogous definitions for $P(X|\alpha,\beta)$ and $P(X,Y|\alpha,\beta)$) with concentration parameter $\alpha$ and base measure hyperparameter $\beta$.
%We employ Bayesian Hierarchical Clustering to approximate the probability of the data under a DPM by computing $P(D|T_k)$ which can be turned into a lower bound on $p(D_k)$ via inequality (\ref{eq:9}).

%The first hypothesis $\Hcal_1$ is that both samples were in fact generated independently and identically from the same probability model $p(x|\theta)$ with unknown parameters $\theta$. To evaluate the probability of the data under this hypothesis we need to specify some prior over the parameters of the model, $P(\theta|\beta)$ with hyperparameters $\beta$. We can then compute the probability of the aggregate sample $D = X \cup Y$ under $\Hcal_1$:
%\begin{align}
%  \label{eq:1}
%  P(D|\Hcal_1) =  \int P(D|\theta) P(\theta|\beta) d\theta.
%\end{align}
%If we choose a model with conjugate priors, this integral is tractable. Examples include the Beta priors for the Binomial distribution or the Dirichlet priors for the Multinomial distribution.
%
%The second hypothesis is that both samples belong to different clusters.
%
%

\subsection{Dirichlet Process Mixture Models}
\label{sec:dirichl-mixt-models}
A key component in our nonparametric two sample test is the ability to approximately infer the marginal probability of a set of observations from a Dirichlet Process Mixture Model (DPM). As these DPMs are at the heart of our nonparametric two-sample test, let us review them here~\cite{Ferguson73,Antoniak74}.

A Dirichlet Process (DP), and also a Dirichlet Process Mixture Model (DPM), is a probability distribution on probability distributions, and 
DPMs consider the limit of infinitely many components of a finite mixture model. By allowing for an infinite number of components, we are able to model the complicated distributions that we encounter in real-world applications via DPMs. 

Consider a finite mixture model with $C$ components
\begin{align}
  \label{eq:2}
  p(x^{(i)}|\phi)= \sum_{j=1}^{C} p(x^{(i)} | \theta_j) p(c_i = j| \zeta)
\end{align}
where $c_i \in \{1,\ldots,C\}$ is a cluster indicator variable for data point $i$, $\zeta$ are the parameters of a multinomial distribution with \begin{align} p(c_i=j|\zeta) = \zeta_j, \end{align} $\theta_j$  are the parameters of the $j$th component, and \begin{align} \phi = (\theta_1,\ldots,\theta_C,\zeta).\end{align} 

Let the parameters of each component have conjugate priors $p(\theta|\beta)$ as before, and the multinomial parameters also have a conjugate Dirichlet prior
\begin{align}
  \label{eq:3}
  p(\zeta| \alpha) = \frac{\Gamma(\alpha)}{\Gamma(\alpha/C)^C} \prod_{j=1}^{C} \zeta_j^{\alpha/C-1}
\end{align}

Given a data set $\Dcal = \{x^{(1)},\ldots, x^{(n)}\}$, the marginal likelihood for this mixture model is 
\begin{align}
  \label{eq:4}
  p(\Dcal|\alpha,\beta) = \int [\prod_{i=1}^{n} p(x^{(i)}| \phi)] p(\phi|\alpha,\beta)d\phi,
\end{align}
where 
\begin{align}p(\phi| \alpha, \beta) =  p(\zeta|\alpha)\prod_{j=1}^{C} p(\theta_j|\beta). \end{align} This marginal likelihood can be rewritten as 
\begin{align}
  \label{eq:5}
  p(\Dcal|\alpha,\beta) = \sum_c p(c|\alpha)p(\Dcal|c,\beta)
\end{align}
where $c= (c_1,\ldots,c_n)$ and \begin{align} p(c|\alpha) = \int p(c|\zeta) p(\zeta|\alpha) d\zeta \end{align} is a standard Dirichlet integral. The quantity~(\ref{eq:5}) is well-defined even in the limit $C \rightarrow \infty$. Although the number of possible settings of $c$ grows as $C^n$ and therefore diverges as $C \rightarrow \infty$, the number of possible ways of partitioning the n points remains finite (roughly $O(n^n)$). Using $\Vcal$ to denote the set of all possible partitioning of $n$ data points, we can re-write (\ref{eq:5}) as 
\begin{align}
  \label{eq:7}
  p(\Dcal|\alpha,\beta) = \sum_{v \in \Vcal} p(v|\alpha) p(\Dcal|v,\beta)
\end{align}
 
\subsection{Approximate inference of marginal probabilities under DPM}
\label{sec:dirichl-proc-mixt}

While finite, the number of partitions still grows as $O(n^n)$ with the size $n$ of the dataset, rendering an exact inference of the marginal probabilities under a DPM intractable even for moderate size datasets (roughly $n > 10$).
Hence we have to resort to approximate inference methods for computing these marginals. One choice is Bayesian hierarchical clustering (BHC), a clustering algorithm that can be used for approximate inference of marginal probabilities under a DPM in $O(n^2)$~\cite{heller_bayesian_2005}. 
%We are making use of this technique when approximating the marginal probabilities of the data under a DPM in our experiments.  

\subsection{Bayes factor in nonparametric test}
\label{sec:bayes-fact-nonp}
For the nonparametric two-sample test we use a DPM as the distribution on distributions $q$, $q_1$, $q_2$. This allows us to integrate out the parameters of the unknown underlying probability distributions $q$, $q_1$ and $q_2$ in a Bayesian manner, while employing a flexible model for these distributions.

The Bayes factor $\chi$ from (\ref{eq:6}) can then be computed as 
\begin{align}
  \label{eq:10}
\chi & =  \frac{\int P(X| q_1) P(q_1|\alpha, \beta) dq_1 * \int P(Y| q_2) P(q_2|\alpha, \beta) dq_2} {\int P(X,Y| q) P(q|\alpha, \beta) d q} \\ & = \frac{ P(X|\alpha, \beta)  P(Y|\alpha, \beta)}{P(X,Y|\alpha, \beta)},\label{eq:13}
\end{align}

where $P(X|q_1)=\prod_{i=1}^{m_1} q_1(x_i)$ and $P(q_1|\alpha,\beta)$ is a Dirichlet process mixture with concentration parameter $\alpha$ and base measure hyperparameter $\beta$. Hence $P(X,Y|\alpha, \beta)$ denotes the marginal probability that $X$ and $Y$ were generated from this DPM with hyperparameters $\alpha$ and $\beta$ (analogous for $P(X|\alpha, \beta)$ and $P(Y|\alpha, \beta)$). %We approximate all three terms in (\ref{eq:13}) by means of Bayesian Hierarchical Clustering~\cite{heller_bayesian_2005} in our experiments.

\section{Discussion and Conclusions}
\label{sec:discussion}

In this paper, we have proposed two classes of Bayesian two sample tests, a parametric test based on distributions from the exponential family, and a nonparametric test based on Dirichlet Process Mixture Models.

% Both on synthetic data and real-world medical datasets, our Bayesian tests prove to be competitive with the best frequentist two-sample tests. When we generate synthetic data according to some distribution from the exponentially family, the parametric test unsurprisingly is more accurate than the nonparametric test. When we are dealing with real-world data that are often generated from complex underlying distributions, the nonparametric test is able to capture these distributions more accurately. 

An issue of future work will be the runtime of two-sample tests. Frequentist tests are often expensive to compute, as the test statistic often requires at least an runtime of $O(n^2)$ for $n$ datapoints and bootstrapping for determining the null distribution. The Bayesian tests avoid this bootstrapping step and there exist various approximations to a Dirichlet process mixture model~\cite{BleJor05,DBLP:conf/nips/KuriharaWV06,DBLP:conf/ijcai/KuriharaWT07}, some of which can be computed in less than $O(n^2)$. Hence the Bayesian approach might hold the key for efficient two-sample tests, which we will look at in future work.

\bibliographystyle{mlapa}
\bibliography{bibfile,zotero,zotero_home}

\end{document}